\documentclass[sigconf]{acmart}
\AtBeginDocument{%
  }

\setcopyright{acmlicensed}
\copyrightyear{2026}
\acmYear{2026}
\acmDOI{XXXXXXX.XXXXXXX} 
\acmConference[WWW '26 Industry Track]{Proceedings of the Web Conference 2026 Industry Track}{April 13--17, 2026}{Dubai, UAE}
\acmISBN{978-1-4503-XXXX-X/2026/04} 

\newcommand{\figref}[1]{Figure~\ref{#1}}
\newcommand{\tabref}[1]{Table~\ref{#1}}

\usepackage{algorithm}
\usepackage{algorithmic}
\usepackage{tabularx}
\usepackage{booktabs}
\usepackage{multirow}

\begin{document}

\title{MedKGEval: A Knowledge Graph-Based Multi-Turn Evaluation Framework for Open-Ended Patient Interactions with Clinical LLMs}

\author{Yuechun Yu}
\authornote{Both authors contributed equally to this research.}
\email{yuyuechun.yyc@antgroup.com}
\author{Han Ying}
\authornotemark[1]
\email{yinghan.yh@antgroup.com}
\affiliation{%
  \institution{Ant Group}
  \city{Hangzhou}
  \state{Zhejiang}
  \country{China}
}

\author{Haoan Jin}
\email{pilgrim@sjtu.edu.cn}
\affiliation{%
  \institution{Shanghai Jiao Tong University}
  \city{Shanghai}
  \country{China}
}

\author{Wenjian Jiang}
\email{jiangwenjian.jwj@antgroup.com}
\author{Dong Xian}
\email{xiandong.xd@antgroup.com}
\affiliation{%
  \institution{Ant Group}
  \city{Hangzhou}
  \state{Zhejiang}
  \country{China}
}

\author{Binghao Wang}
\email{binghao.wbh@antgroup.com}
\author{Zhou Yang}
\email{fenghan.yz@antgroup.com}
\affiliation{%
  \institution{Ant Group}
  \city{Hangzhou}
  \state{Zhejiang}
  \country{China}
}

\author{Mengyue Wu}
\authornote{Corresponding author.}
\email{mengyuewu@sjtu.edu.cn}
\authornotemark[2]
\affiliation{%
  \institution{Shanghai Jiao Tong University}
  \city{Shanghai}
  \country{China}
}

\renewcommand{\shortauthors}{Yu et al.}

\begin{abstract}
The reliable evaluation of large language models (LLMs) in medical applications remains an open challenge, particularly in capturing the complexity of multi-turn doctor-patient interactions that unfold in real clinical environments. Existing evaluation methods typically rely on post hoc review of full conversation transcripts, thereby neglecting the dynamic, context-sensitive nature of medical dialogues and the evolving informational needs of patients. In this work, we present \textbf{MedKGEval}, a novel multi-turn evaluation framework for clinical LLMs grounded in structured medical knowledge. Our approach introduces three key contributions: (1) \textbf{a knowledge graph-driven patient simulation mechanism}, where a dedicated control module retrieves relevant medical facts from a curated knowledge graph, thereby endowing the patient agent with human-like and realistic conversational behavior. This knowledge graph is constructed by integrating open-source resources with additional triples extracted from expert-annotated datasets; (2) \textbf{an in-situ, turn-level evaluation framework}, where each model response is assessed by a Judge Agent for clinical appropriateness, factual correctness, and safety as the dialogue progresses—rather than retrospectively—using a suite of fine-grained, task-specific metrics; (3) \textbf{a comprehensive multi-turn benchmark of eight state-of-the-art LLMs}, demonstrating MedKGEval's ability to identify subtle behavioral flaws and safety risks that are often overlooked by conventional evaluation pipelines. Although initially designed for Chinese and English medical applications, our framework can be readily extended to additional languages by switching the input knowledge graphs, ensuring seamless bilingual support and domain-specific applicability. \textbf{All code and datasets of this work are available from the authors upon request.}
\end{abstract}


\begin{teaserfigure}
    \centering
    \includegraphics[width=0.9\linewidth]{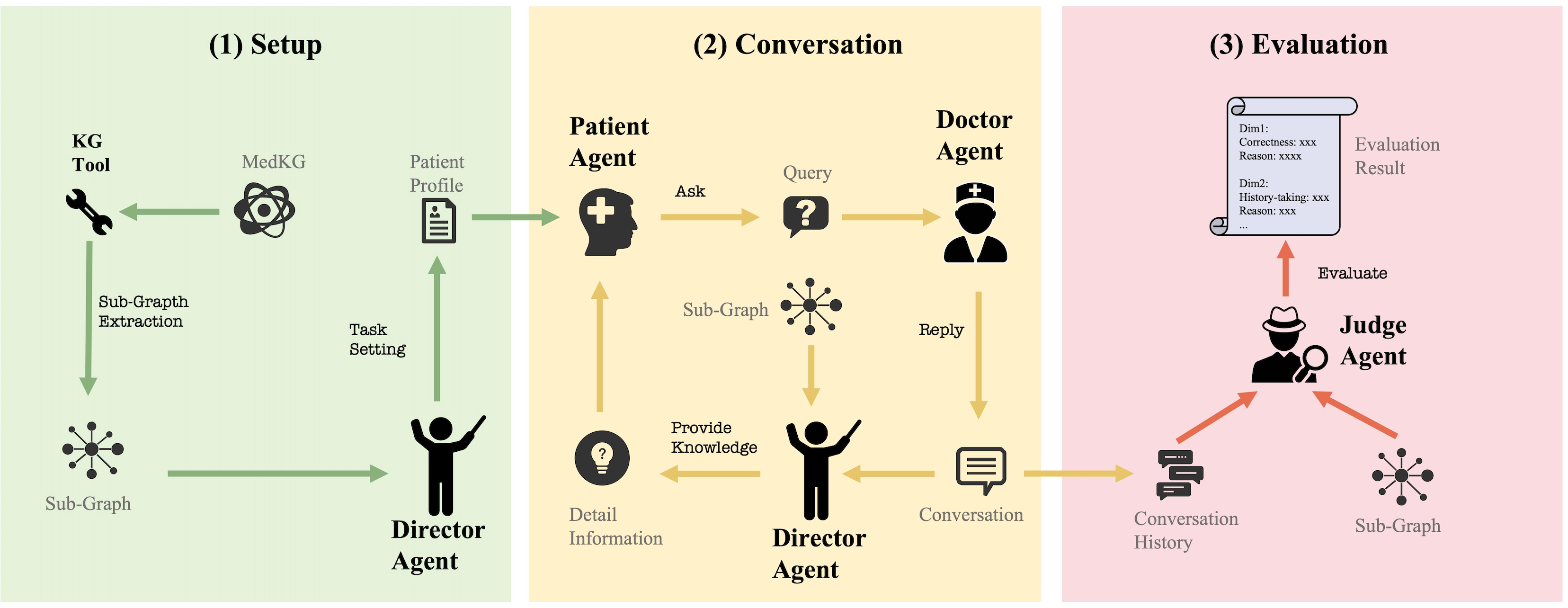}
    \caption{Overview of the MedKGEval pipeline. The framework simulates doctor-patient interactions based on the knowledge graph. A director agent is engaged to promise the patient agent’s stable and consistent behavior in communication. The judge agent enables comperhensive assesment of the LLM's performance on multi-turn conversation.}
    \label{fig:pipeline}
\end{teaserfigure}


\maketitle

\section{Introduction}

Large language models (LLMs) have demonstrated impressive capabilities across a range of natural language processing tasks, including medical question answering, diagnosis support, and patient triage~\citep{brown2020language, singhal2023large}. Their ability to generate fluent, context-aware responses makes them promising tools for clinical decision support and patient-facing applications~\citep{spotnitz2024survey, wang2023large}. However, the reliable evaluation of LLMs in healthcare remains an open challenge—particularly when it comes to assessing performance in open-ended, multi-turn interactions that mirror real-world doctor-patient conversations.

Such conversations often evolve over several turns, with each exchange informed by prior context and shaped by the information newly revealed by the patient. In a typical medical consultation, for instance, a patient might begin by describing a symptom (“I’ve had a persistent cough for two weeks”), the physician then asks a clarifying question (“Have you experienced any fever or shortness of breath?”), the patient adds further detail (“Yes, I developed a mild fever yesterday”), and the dialogue continues until sufficient data is gathered for clinical reasoning. This inherently open-ended trajectory and dependency on earlier turns not only increase the interaction’s complexity but also pose unique challenges for evaluation.
Existing evaluation paradigms typically rely on static benchmarks~\citep{liu2024medbench, arora2025healthbench} or post-hoc analysis of conversation transcripts~\citep{jeblick2024chatgpt, singhal2025toward}, which fail to capture the dynamic and evolving nature of medical dialogue. These approaches overlook how clinical relevance, factual accuracy, and safety may fluctuate across conversational turns. Moreover, without fine-grained, real-time assessment, problems such as error propagation and context drift that emerge over successive turns may go unnoticed.

Recent work has attempted to address these gaps by introducing interactive patient simulators~\citep{liao2024automatic, johri2025evaluation} and incorporating medical knowledge into evaluation~\citep{zuo2025kg4diagnosis}. However, these efforts often still perform evaluation retrospectively and lack systematic integration of domain knowledge at each conversational turn. 

To address these challenges, we present \textbf{MedKGEval}—a knowledge graph-driven \emph{multi-agent} evaluation framework designed to simulate realistic, dynamic medical dialogues and assess LLM behavior in real time. As shown in~\figref{fig:pipeline}, MedKGEval models four key roles: 
(i) a \emph{Doctor Agent} (the LLM under evaluation) that converses with patients and provides diagnoses or medication recommendations; 
(ii) a \emph{Patient Agent} that follows a predefined persona, uses disease or medication information to respond naturally; 
(iii) a \emph{Judge Agent} that evaluates each doctor response for clinical appropriateness and factual accuracy at every turn; 
and (iv) a \emph{Director Agent} that initializes patient personas, resolves conflicting symptoms, and supplies the patient with essential details reflecting realistic clinical constraints. 
The knowledge graph underpins all agents, ensuring that medical facts and symptom–disease–medication relationships are consistently available during conversation.

By combining these roles, MedKGEval enables \emph{fine-grained, turn-by-turn evaluation} that reflects how accuracy and relevance change as dialogues progress—effectively addressing the open-ended and multi-turn nature of clinical interactions. This design overcomes the retrospective limitation of prior benchmarks by embedding evaluation directly into the interaction loop, allowing early detection of context drift, error propagation, and safety breaches.



\begin{figure*}[ht]
    \centering
    \includegraphics[width=0.95\linewidth]{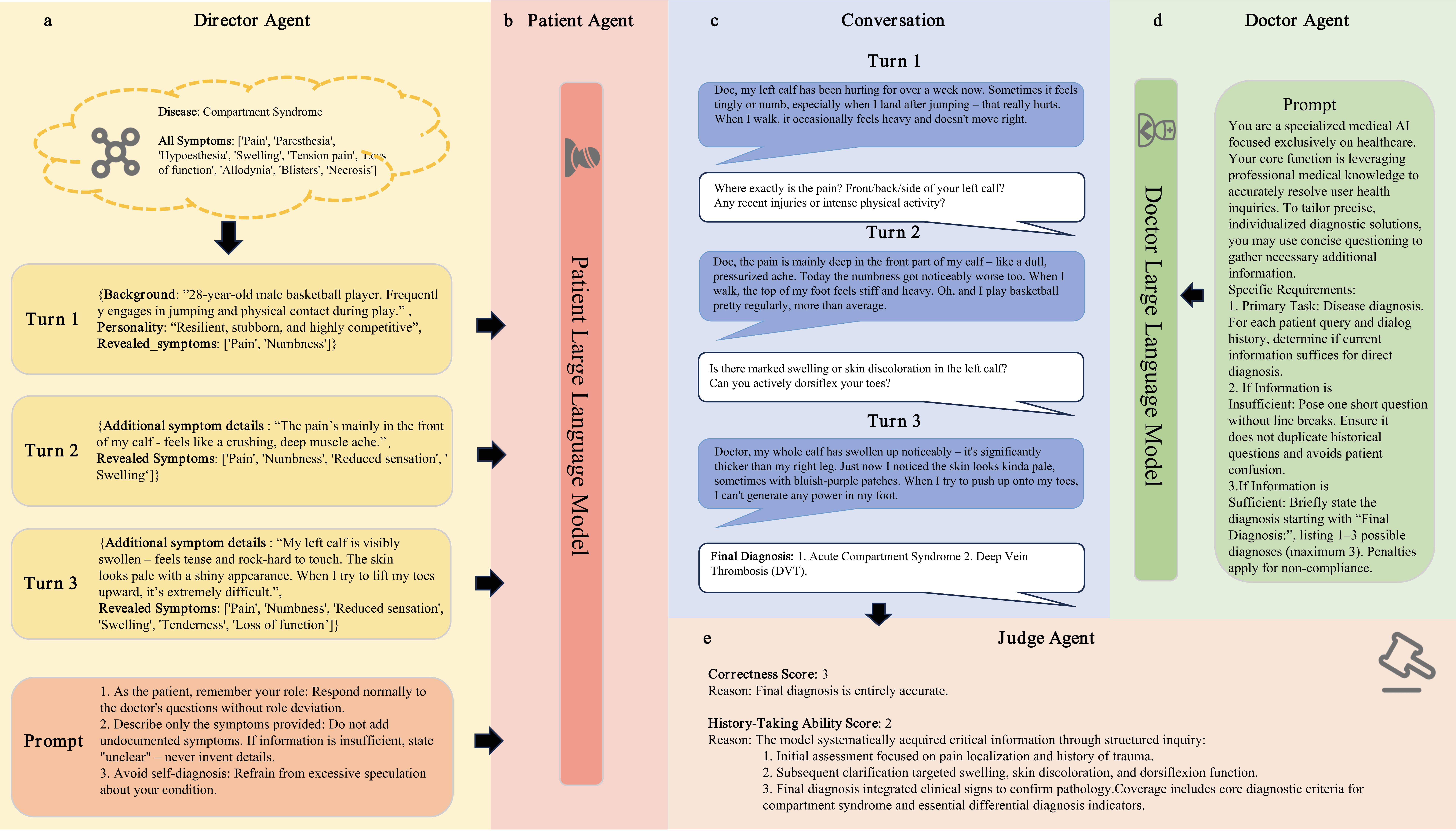}
    \caption{An example case of MedKGEval in the disease-diagnosis task.}
    \label{fig:case}
\end{figure*}

To validate our approach, we focus on two representative medical scenarios—medication consultation and disease diagnosis—and benchmark the performance of both general-purpose and domain-specific LLMs. An illustrative disease-diagnosis case is presented in \figref{fig:case}. The experimental results demonstrate that MedKGEval identifies subtle behavioral flaws and clinical oversights that are often missed by traditional evaluation methods.

Our contributions can be summarized as follows:
\begin{itemize}
    \item We propose a novel evaluation framework for clinical LLMs based on the knowledge graph we constructed, enabling realistic and scalable multi-turn testing.
    \item We construct a Judge Agent assesses each model response for clinical appropriateness, factual accuracy, and safety in real-time using task-specific metrics.
    \item We empirically demonstrate that our framework uncovers clinically relevant failure modes across multiple model types and tasks.
\end{itemize}

\section{Related Work}

\paragraph{Evaluation Benchmarks for Medical LLMs.}  
Benchmarks such as MedBench~\citep{liu2024medbench} and HealthBench~\citep{arora2025healthbench} focus on single-turn question answering or summarization tasks using static datasets. While useful for broad coverage, they fall short in evaluating the longitudinal reasoning and safety awareness required in real clinical conversations.Furthermore, closed evaluations based on fixed datasets risk triggering an "overfitting during evaluation" phenomenon—models may achieve artificially inflated scores by memorizing high-frequency question patterns, yet significantly underperform in actual open-domain conversational scenarios.

\paragraph{Knowledge Graph Applications in Medical LLMs.}
Emerging research demonstrates how knowledge graphs (KGs) address critical limitations of LLMs in clinical applications. \citet{yang-etal-2024-kg} proposed KG-Rank, which integrates medical KGs with ranking-enhanced triplet retrieval to improve clinical QA accuracy. For complex decision-making tasks, \citet{jiang2025reasoningenhanced} developed KARE, a framework that combines hierarchical KG community detection with LLM reasoning, reducing clinical prediction hallucinations through context-aware knowledge aggregation. However, existing KG applications primarily focus on model training or clinical decision support, leaving under-explored their potential in constructing evaluation benchmarks. KGs are well-suited for benchmarks: their structured, composable relations (e.g., diagnoses, treatments, adverse events, temporal and lab findings) and multi-hop paths enable gold-standard reasoning chains and assessment of procedural reasoning, path consistency, and factual faithfulness. Thus, KGs offer an interpretable, controllable, and scalable basis for benchmark construction, addressing gaps in content breadth and process-aware evaluation.


\paragraph{Multi-Turn and Interactive Evaluation.}  
Several frameworks have been proposed to move toward interactive evaluation, including MT-Eval~\citep{kwan2024mt} and MultiChalleng~\citep{sirdeshmukh2025multichallenge}. These efforts evaluate multi-turn conversation as a whole, thus fails to provide more fine-grained evaluation result up to each turn. \citet{liao2024automatic} reformulates medical multiple-choice questions into patient information and diagnostic tasks. However, their work is limited by the finite nature of the datasets derived from medical exams, which restricts the scalability of the evaluation scenarios. In contrast, our framework leverages a knowledge graph-based framework, allowing for the dynamic and virtually limitless expansion of consultation tasks. This graph-based pipeline not only enhances the diversity of the evaluation but also ensures that the system can be readily updated with new medical knowledge. Furthermore, under the guidance of a dedicated control module, our Patient Agent demonstrates stable and consistent interactions, maintaining coherence in communication throughout the consultation process.

\section{Dataset Construction}
\label{sec:dataset-construction}

To support structured and clinically grounded evaluation of medical LLMs across both Chinese and English settings, we construct task-specific medical knowledge graphs, which we collectively term \textbf{MedKG}. These graphs are used to systematically generate multi-turn test cases for our evaluation scenarios. Specifically, the Chinese version of MedKG is built upon CMeKG~\citep{byambasuren2019preliminary}, an open-source medical KG, and the English version is built upon PrimeKG~\citep{chandak2023building}; both are selectively curated and extended to focus on entities and relations most relevant to our two core evaluation tasks: \emph{disease diagnosis} and \emph{medication consultation}.

Specifically, we extract a disease-centered subgraph by identifying all disease nodes and retrieving their directly associated \textit{symptoms} and \textit{drugs}. For each drug entity, we further incorporate key clinical attributes such as \textit{indications}, \textit{contraindications}, \textit{precautions}, and \textit{drug-drug interactions}, capturing the contextual factors critical to appropriate drug usage. To improve domain coverage and factual reliability, we augment the initial graph with additional triples extracted from high-quality medical QA corpora, including MedQA~\citep{jin2021disease} and NLPEC~\citep{li-etal-2020-towards}, emphasizing pharmacological relations with high semantic confidence.

In practice, medical knowledge graphs often aggregate heterogeneous sources—including clinical statistics, historical case archives, and standardized medical textbooks—which may introduce biologically inconsistent or clinically implausible associations~\citep{wu2023medical}. For example, naive integration can result in symptom clusters that violate basic medical logic, such as male patients presenting with ovarian cysts. 

\begin{figure}[ht]
    \centering
    \includegraphics[width=0.9\linewidth]{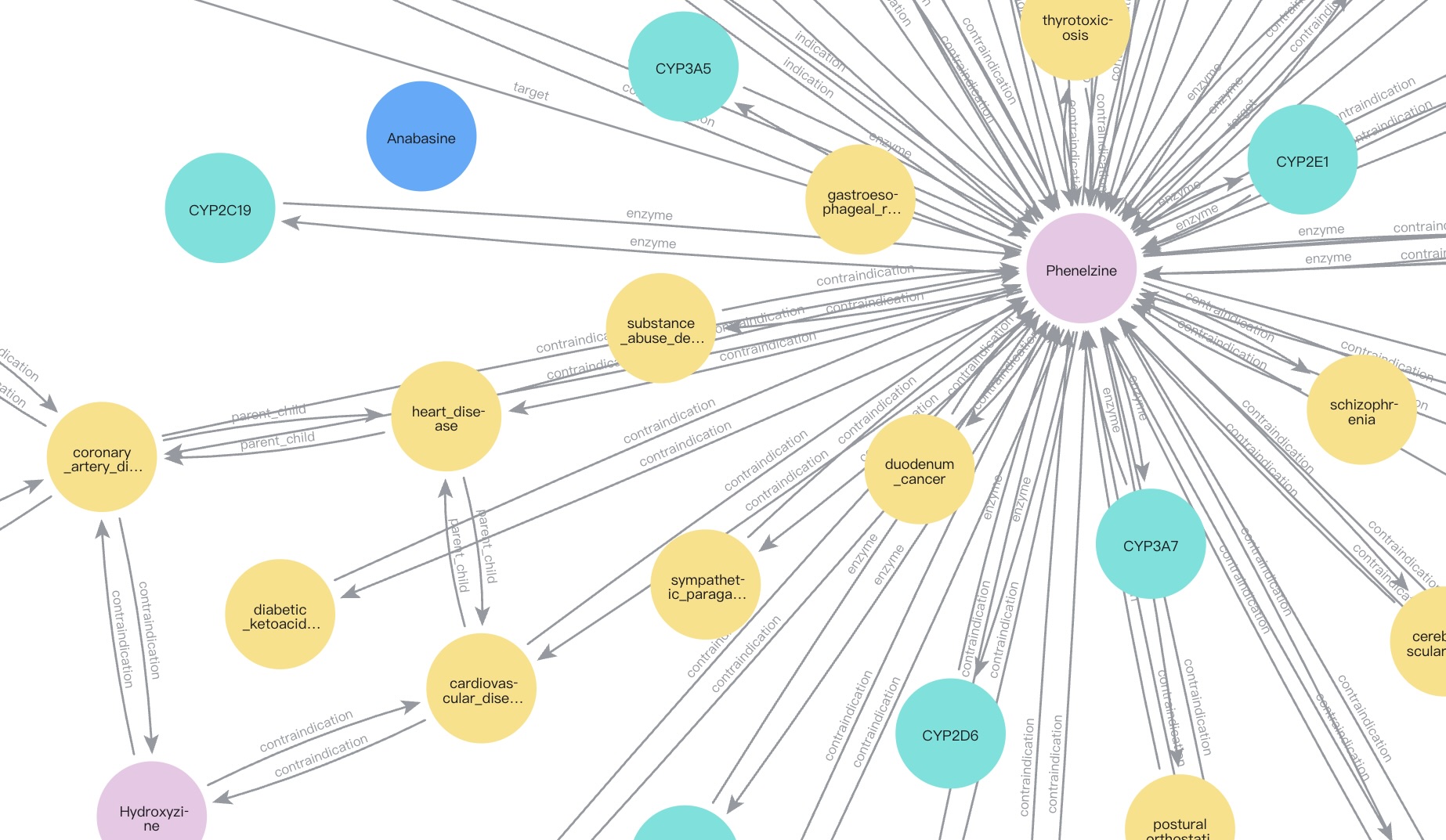}
    \caption{Illustration of the curated medical knowledge graph used for evaluation.}
    \label{fig:kg-construction}
\end{figure}

To mitigate such noise and enhance the clinical realism of simulated interactions, we design a domain-specific cleaning process consisting of:  
(1) \emph{semantic–consistency filtering}, which removes medically invalid triples that conflict with basic constraints such as gender, anatomy, or pathophysiology; and  
(2) \emph{discriminative–symptom selection}, which prioritizes diagnostically distinctive features (e.g., Koplik spots for measles, resting tremor for Parkinson's disease). The resulting graph thus provides a high-quality, clinically coherent foundation for downstream case generation and evaluation, ensuring that medical facts and symptom–disease–medication relationships are consistently available during doctor-aptient conversations. An illustration of our graph is shown in \figref{fig:kg-construction}, and \tabref{tab:kg-all-stats} presents its statistical overview.


\begin{table}[htpb]
\centering
\begin{tabular}{lccc}
\toprule
\textbf{Category}          & \textbf{Count} \\
\midrule
Relations                 & 651,463 \\
Drugs                     & 15,164 \\
Diseases                  & 15,898 \\
Symptoms                  & 32,248 \\
Indications          & 54,203 \\
Complication         & 31,924 \\
Adverse Reactions    & 165,976 \\
Precautions          & 26,414 \\
\bottomrule
\end{tabular}
\caption{Statistics of entities in the knowledge graph.}
\label{tab:kg-all-stats}
\end{table}

\section{Pipeline Construction}
\label{sec:pipeline}

Addressing the lack of standardized evaluation for medical conversational reasoning, we introduce \textbf{MedKGEval}, a modular and agent-driven pipeline that leverages the medical knowledge graph to conduct rigorous, multi-turn evaluations of LLMs. An overview of the complete framework is illustrated in \figref{fig:pipeline}. 

To capture complementary aspects of clinical reasoning, we design two representative evaluation scenarios :
\begin{itemize}
    \item \textbf{Medication Consultation:} Given a target drug, we first retrieve its structured attributes from the knowledge graph (e.g., indications, contraindications, precautions), which are then used to guide the behavior of the patient simulator. Specifically, the simulation proceeds through multiple dialogue rounds, each corresponding to a different attribute information. For instance, in the first round, the patient may ask whether the drug is suitable for a condition listed under its indications. In the next round, they may pose a question that implicitly tests knowledge of a contraindication (e.g., “I have high blood pressure—can I still take this?”). Subsequent rounds focus on precautionary factors, such as pregnancy or pediatric use. To enhance realism, each simulated query is conditioned on a dynamically generated patient persona, ensuring that the dialogue reflects plausible real-world inquiry patterns. At each turn of the conversation, a judge module evaluates the LLM’s response along two key dimensions: accuracy and comprehensiveness.
    \item \textbf{Disease Diagnosis:} For a given target disease, we first extract its associated symptoms from the knowledge graph and filter out those that are conflicting or irrelevant.  A simulated patient then presents the remaining symptoms incrementally across several dialogue rounds, each time revealing one or two findings while adhering to a predefined persona and narrative style.  When the LLM seeks clarification on specific symptoms, the simulator retrieves corresponding details from the knowledge graph if available; otherwise, it generates contextually appropriate filler responses. The dialogue continues until the LLM offers a final diagnosis or until all symptoms have been presented—at which point the patient explicitly requests a diagnosis. The model’s performance is evaluated through a dual scoring mechanism: the final diagnosis is assessed for correctness, while intermediate turns are rated based on the quality of medical history-taking.
\end{itemize}
These scenarios allow MedKGEval to evaluate both knowledge application in medication-related decision-making and reasoning in diagnostic settings.

To implement these scenarios, \textbf{MedKGEval} consists of four specialized agents:

\paragraph{Doctor Agent.}  
This is the primary evaluation target—the medical LLM under assessment. It receives patient inputs turn by turn and responds accordingly, simulating the role of a clinician. The doctor agent remains stateless across different test cases but accumulates dialogue history within each session to enable context-aware reasoning.

\paragraph{Director Agent.}  
As noted in Section \textbf{Dataset Construction}, the Director Agent serves as a centralized controller. It retrieves relevant information from the knowledge graph and filters the extracted subgraph nodes according to medical rationale to construct clinically plausible patient profiles. The agent then dynamically determines the next intent or query to guide the simulated patient: for drug-related tasks, it sequentially reveals attributes such as contraindications or precautions; for diagnostic tasks, it modulates the pace of symptom disclosure and controls the dialogue length. When the Doctor Agent inquires about additional details, the Director Agent supplies disease-specific guidance to the Patient Agent, preventing responses from straying from realistic clinical presentations and thereby preserving the diagnostic validity of the disclosed symptoms.

\paragraph{Patient Agent.}  
Guided by the Director Agent’s instructions, the Patient Agent produces natural-language utterances that mimic real patient behavior. It faithfully follows the assigned persona—age, medical history, and communication style—and introduces symptom descriptions or medication-related questions in a coherent, human-like manner. Crucially, the Patient Agent begins with limited medical knowledge, reflecting real-world clinical encounters where patients often struggle to articulate precise symptoms or identify relevant details during initial consultations. The Director Agent therefore unveils critical medical information gradually, timing each disclosure to match the natural flow of a physician’s diagnostic questioning.

\paragraph{Judge Agent.}  
Our Judge Agent, leveraging multi-turn dialogue history and relevant medical knowledge provided by the God Agent, assesses model behavior along two tasks: \textit{Medication Consultation} and \textit{Disease Diagnosis}. For each tasks, we move beyond binary correctness and introduce multi-level scoring frameworks that reflect the nuanced requirements of clinical decision-making.

For \textbf{Medication Consultation}, we focus on two core criteria: 
(1) \textbf{correctness}—whether the model’s response aligns with established pharmacological knowledge and avoids harmful misinformation; and 
(2) \textbf{comprehensiveness}—whether the model delivers a thorough and informative response addressing all key aspects of the user’s query. Both dimensions are critical for ensuring clinical safety and the practical utility of medication-related advice, as incomplete guidance can lead to suboptimal or unsafe treatment outcomes.

For \textbf{Disease Diagnosis}, we assess two key dimensions: 
(1) \textbf{correctness}—whether the model's final prediction aligns semantically or ontologically with the reference disease; and 
(2) \textbf{history-taking skill}—whether the model effectively gathers relevant clinical information across turns to support accurate differential diagnosis.

Notably, the history-taking metric is a novel addition to automated clinical LLM evaluation. Unlike prior benchmarks focusing solely on the final diagnosis, our framework assesses the incremental gathering and synthesis of symptoms and background information—a cornerstone of safe and effective clinical reasoning~\citep{johri2025evaluation}. This skill is vital in multi-turn dialogues, where each follow-up question can shape diagnostic accuracy. Our in-situ scoring protocol captures this process turn-by-turn, enabling fine-grained detection of reasoning gaps. The history-taking evaluation is therefore both clinically grounded and methodologically distinct to our framework.

The detailed scoring rubric is summarized in \tabref{tab:judge-scoring}.

\begin{table}[ht]
\centering
\small
\renewcommand{\arraystretch}{1.2}
\setlength{\tabcolsep}{4pt}
\begin{tabularx}{\linewidth}{l|c|X}
\toprule
\textbf{Task} & \textbf{Score} & \textbf{Criteria Description} \\
\midrule
\multirow{3}{*}{\shortstack[l]{\textbf{Drug}\\Correctness}} 
& 0 & All key facts incorrect; content contradicts medical knowledge. \\
& 1 & Mixed correctness; some key facts incorrect, others accurate. \\
& 2 & Fully accurate; all key facts align with reference. \\
\midrule
\multirow{3}{*}{\shortstack[l]{\textbf{Drug}\\Comprehensiveness}} 
& 0 & Omits all relevant information. \\
& 1 & Includes partial relevant info; misses minor but meaningful points. \\
& 2 & Fully covers all user-query-relevant information. \\
\midrule
\multirow{4}{*}{\shortstack[l]{\textbf{Disease}\\Correctness}} 
& 0 & Diagnosis is clinically unrelated to ground truth (no symptom, cause, or taxonomy overlap). \\
& 1 & Clinically similar but no ontological relation (e.g., symptom overlap but different categories). \\
& 2 & Ontologically related (e.g., parent-child disease, shared mechanism). \\
& 3 & Exact match or clinical synonym. \\
\midrule
\multirow{3}{*}{\shortstack[l]{\textbf{Disease}\\History-taking}} 
& 0 & Collected info is irrelevant; cannot support any differential diagnosis. \\
& 1 & Partial core info collected; misses discriminative/exclusionary features. \\
& 2 & Structured inquiry; covers key disease-specific features. \\
\bottomrule
\end{tabularx}
\caption{Scoring rubric for Judge Agent across Medication Consultation and Disease Diagnosis.}
\label{tab:judge-scoring}
\end{table}

\section{Experiments}
\label{sec:experiments}

In our evaluation setup, the primary objective is to rigorously assess the capabilities and safety of the \textbf{Doctor Agent}. Our framework incorporates benchmark testing with state-of-the-art models including Qwen3, GPT-4o, LLaMA3.1, DeepSeek-R1 (general-purpose) and Lingshu, MedGemma, Citrus, Huatuo (medical-specialized). The evaluation tasks focus on two clinical dimensions:
\begin{itemize}
\item Medication Consultation: Evaluated on 500 Chinese and 500 English subgraphs derived from \textbf{MedKG}, centered on medications with nodes for drug names, indications, contraindications, and precautions.
\item Disease Diagnosis: Evaluated on 500 Chinese and 500 English subgraphs derived from \textbf{MedKG}, centered on diseases with nodes describing pathological entities, clinical symptoms, and comorbid conditions.
\end{itemize}
To enable this, our pipeline first requires the careful selection of three supporting agents—\emph{Director Agent}, \emph{Patient Agent}, and \emph{Judge Agent}—each serving a distinct functional role within the evaluation process.

\subsection{Agent Configuration}

\paragraph{Director Agent.}
As the orchestrator of the dialogue environment, the Director Agent must possess strong general reasoning ability, robust instruction-following behavior, and broad domain knowledge across both Chinese and English. To this end, we adopt DeepSeek-R1-671B~\citep{guo2025deepseek}, which demonstrates state-of-the-art performance in Chinese language understanding and generation, while also maintaining highly competitive capabilities in English. This makes it well-suited for multilingual dialogue scenarios, including complex domains such as medicine.

\paragraph{Judge Agent.}
The Judge Agent is responsible for assessing the correctness and comprehensiveness of each Doctor Agent's responses. Since our tasks are defined by a powerful Director with standardized answers, the evaluation architecture can adopt a more lightweight model paradigm that focuses on rule comprehension and reference knowledge validation, rather than open-ended correctness verification. In this closed-domain setting, we prioritize strong bilingual (Chinese and English) language competency alongside efficient validation capabilities when selecting models. We therefore conduct a human-in-the-loop agreement test using several large language models that demonstrate high performance in both Chinese and English, including QwQ-32B~\citep{yang2024qwen2}, Qwen3-32B~\citep{yang2025qwen3}, and DeepSeek-R1-32B~\citep{guo2025deepseek}. Our preliminary results show that QwQ-32B achieves the highest agreement with expert human raters. Detailed agreement statistics are presented in \tabref{tab:human-alignment}.
\begin{table}[ht]
\centering

\begin{tabular}{lccccc}
\toprule
\textbf{Model} & \multicolumn{2}{c}{\textbf{MC}} & \multicolumn{2}{c}{\textbf{DD}}  \\
\cmidrule(lr){2-3} \cmidrule(lr){4-5}
& \textbf{Corr} & \textbf{Comp} & \textbf{Corr} & \textbf{HT} &  \\
\midrule
QwQ-32B & \textbf{0.1370} & \textbf{0.2911} & \textbf{0.3569} & \textbf{0.2815}  \\
Qwen3-32B & 0.2204 & 0.3871 & 0.6316 & 0.5081  \\
DeepSeek-R1-32B & 0.2173 & 0.3858 & 0.6361 & 0.4783 \\
\bottomrule
\end{tabular}
\caption{Mean Absolute Error (MAE) between Judge Agent ratings and human ratings across different metrics for Medication Consultation(\textbf{MC}) and Disease Diagnosis(\textbf{DD}) tasks. \textbf{Note:} ``\textbf{Corr}'' denotes Correctness, ``\textbf{Comp}'' denotes Comprehensiveness and ``\textbf{HT}'' denotes History-taking.}
\label{tab:human-alignment}
\end{table}

\paragraph{Patient Agent.}
To minimize noise and maintain consistency across evaluations, the Patient Agent must exhibit stable and controlled behavior. Thus we compare three candidate models—QwQ-32B, Qwen3-32B, and DeepSeek-R1-32B—in terms of their dialogue querys consistency when coummunicate with different Doctor Agents. Using BGE-M3\citep{chen2024bge}, we compute the average embedding similarity across querys sets to quantify intra-agent variability. Specifically, the consistency score (Cons) is calculated as follows:

\begin{equation}
\mathrm{Cons} = \frac{1}{|D|} \sum_{d \in D} \left( \frac{2}{n_d(n_d - 1)} \sum_{1 \leq i < j \leq n_d} \mathrm{sim}(R_{d,i}, R_{d,j}) \right)
\label{eq:consistency}
\end{equation}
where \( D \) denotes the set of diseases, \( n_d \) is the number of patient querys for disease \( d \), and \( \mathrm{sim}(R_{d,i}, R_{d,j}) \) denotes the similarity between the \( i \)-th and \( j \)-th patient's querys for disease \( d \). This metric thus first computes the average pairwise similarity of patient's querys under each disease and then averages across all diseases.

The results show that QwQ-32B achieves the highest consistency with an average similarity score of \textbf{83.86}, outperforming Qwen3-32B (\textbf{81.95}) and DeepSeek-R1-32B (\textbf{81.34}). This superior stability leads us to select QwQ-32B as the final Patient Agent. The specific consistency scores and detailed statistics are shown in \tabref{tab:human-consistency}.

\begin{table}[ht]
\centering

\begin{tabular}{lccc}
\toprule
\textbf{Model} & \textbf{MC} & \textbf{DD} & \textbf{Avg.} \\
\midrule
QwQ-32B & \textbf{84.18\%} & \textbf{83.53\%} & \textbf{83.86\%} \\
Qwen3-32B & 82.06\% & 81.84\% & 81.95\% \\
DeepSeek-R1-32B & 81.79\% & 80.88\% & 81.34\% \\
\bottomrule
\end{tabular}
\caption{Consistency Scores of the Patient Agent in Medication Consultation(\textbf{MC}) and Disease Diagnosis(\textbf{DD}) tasks.}

\label{tab:human-consistency}
\end{table}



\begin{table*}[ht]
\centering
\begin{tabular}{llccccccccc}
\toprule
\textbf{Task} & \textbf{Metric} & \textbf{Language} & \textbf{Qwen3} & \textbf{GPT-4o} & \textbf{LLaMA3.1} & \textbf{DeepSeek} & \textbf{Citrus} & \textbf{Lingshu} & \textbf{Huatuo} & \textbf{MedGemma} \\
\midrule
\multirow{4}{*}{MC} 
& Corr(multi-turn) & EN & 0.7642 & 0.6267 & 0.7408 & \textbf{0.8763} & 0.6911 & 0.5911 & 0.5833 & 0.6606 \\
& Comp(multi-turn)  & EN & 0.6580 & 0.5701 & 0.6099 & \textbf{0.8351} & 0.5733 & 0.4609 & 0.5182 & 0.5674 \\
& Corr(multi-turn) & ZH & 0.8497 & 0.8330 & 0.7300 & \textbf{0.8858} & 0.8534 & 0.8270 & 0.8548 & 0.6697 \\
& Comp(multi-turn)  & ZH & 0.6837 & 0.6406 & 0.6278 & \textbf{0.7581} & 0.6850 & 0.6685 & 0.7021 & 0.5442 \\
\midrule
\multirow{6}{*}{DD} 
& Corr(multi-turn) & EN & 0.4733 & 0.4572 & \textbf{0.4796} & 0.4471 & 0.4274 & 0.4352 & 0.4467 & 0.3991 \\
& HT(multi-turn)  & EN & 0.3000 & 0.4844 & \textbf{0.6973} & 0.2821 & 0.4983 & 0.3488 & 0.4689 & 0.3080 \\
& Corr(sigle-turn) & EN  & 0.4828 & \textbf{0.4879} & 0.4408 & 0.4739 & 0.4717 & 0.4212 & 0.4411 & 0.4358 \\
& Corr(multi-turn) & ZH  & 0.3985 & 0.4375 & 0.4321 & \textbf{0.4888} & 0.3890 & 0.3677 & 0.3875 & 0.3333 \\
& HT(multi-turn)  & ZH & 0.2763 & 0.5000 & 0.6314 & \textbf{0.6894} & 0.3175 & 0.4510 & 0.5670 & 0.3463 \\
& Corr(sigle-turn) & ZH  & 0.4627 & 0.4751 & 0.4384 & \textbf{0.4974} & 0.4703 & 0.3985 & 0.4639 & 0.4490 \\
\bottomrule
\end{tabular}
\caption{General Performance of eight LLMs on Medication Consultation(\textbf{MC}) and Disease Diagnosis(\textbf{DD}) tasks in English (EN) and Chinese (ZH) scenario. \textbf{Note:} ``\textbf{Corr}'' denotes correctness, ``\textbf{Comp}'' denotes comprehensiveness and ``\textbf{HT}'' denotes history-taking.}

\label{tab:task-results}
\end{table*}


\subsection{Results on evaluating MedLLMs}


We evaluate eight state-of-the-art language models as Doctor Agents, focusing on their multilingual performance in Chinese and English across two clinical tasks: Medication Consultation and Disease Diagnosis. Each model is assessed using our scoring rubric and multi-agent evaluation pipeline (see Appendix~\textbf{Model Description} for model details).

\paragraph{Medication Consultation.} 

\tabref{tab:task-results} shows normalized scores (0–1) for multi-turn Medication Consultation, measuring \textit{correctness} and \textit{comprehensiveness}. General-purpose large models consistently outperform smaller medical-specific ones. DeepSeek-R1-671B leads in both languages, with correctness scores of 0.8858 (ZH) / 0.8763 (EN) and comprehensiveness scores of 0.7581 / 0.8351, demonstrating robust cross-lingual medical knowledge. Qwen3-235B-A22B and Citrus-72B follow closely in Chinese, while HuatuoGPT-72B—the top medical-specialized model—matches general models in Chinese but drops significantly in English (correctness: 0.5833 vs. 0.8548).

MedGemma-27B lags in both metrics, indicating that domain specialization alone is insufficient without adequate model capacity. Across all models, \textit{correctness} exceeds \textit{comprehensiveness}, revealing a tendency to provide factually valid but incomplete guidance—a critical safety concern.

\begin{figure*}[ht]
    \centering
    \includegraphics[width=0.8\textwidth]{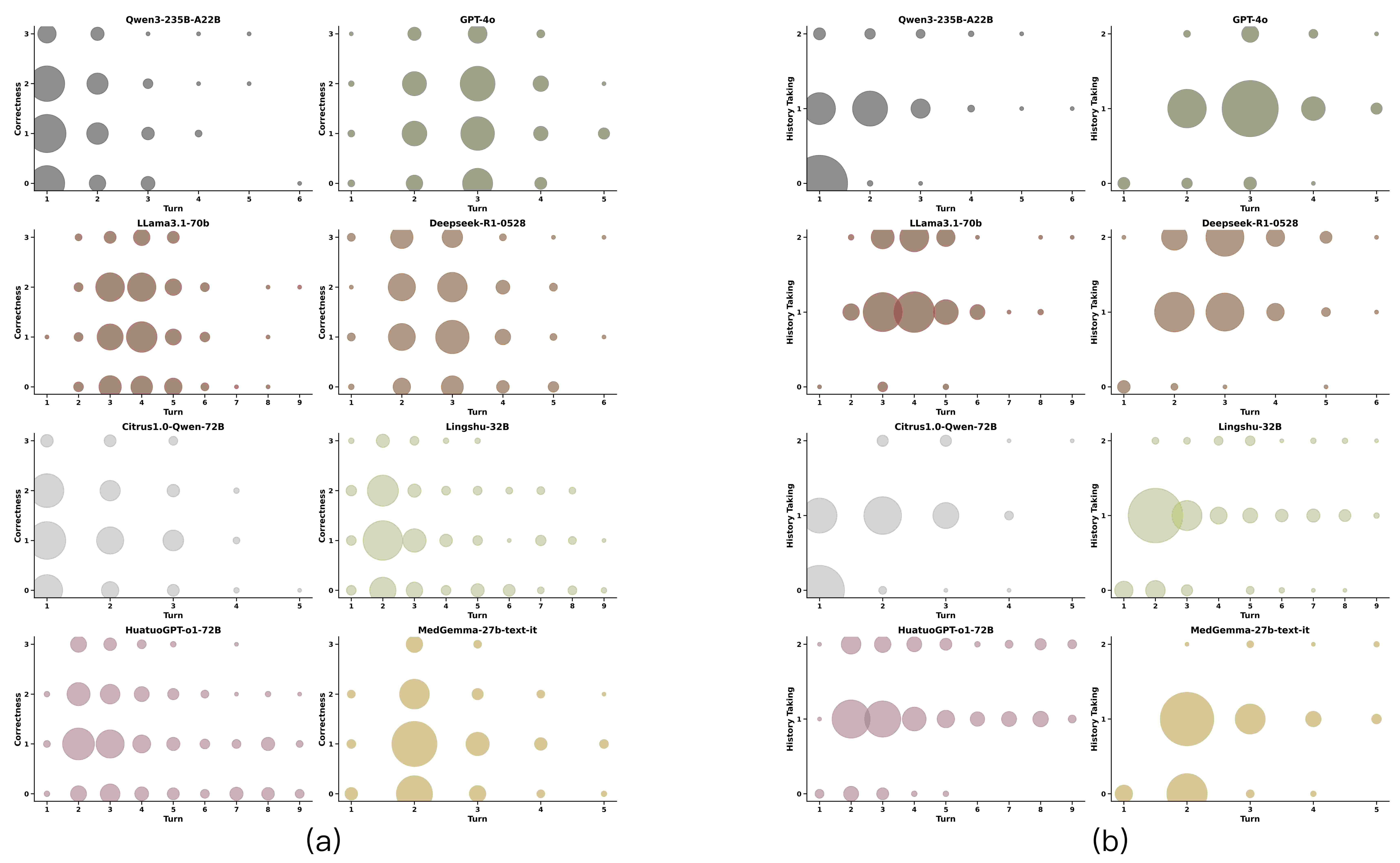}
    \caption{Correctness (a) and History taking scores (b) of eight models in a Chinese disease-diagnosis scenario at different turn counts. Larger circles indicate more cases that ended the dialogue at that turn and received the corresponding score.}
    \label{fig:correctness-turn}
    \label{fig:history-turn}
\end{figure*}

\paragraph{Disease Diagnosis.}

Results are more nuanced. In Chinese, DeepSeek-R1-671B again leads in both multi-turn correctness and history-taking, though most models show a wider gap between history elicitation and final diagnosis. For example, LLaMA3.1-70B excels at history-taking (0.6314) but underperforms in diagnostic correctness, suggesting strong conversational scaffolding but weak diagnostic synthesis. Domain-specific models like HuatuoGPT and Lingshu similarly prioritize symptom recall over end-to-end reasoning. GPT-4o maintains balanced performance, indicating better diagnostic calibration.

In English, rankings shift: LLaMA3.1-70B achieves the highest multi-turn correctness (0.4796) and history-taking (0.6973), while GPT-4o leads in single-turn diagnosis (0.4879). DeepSeek-R1-671B underperforms in English, with particularly low history-taking quality (0.2821). Interaction-level analysis reveals it issues a diagnosis after only 1.11 turns on average in English multi-turn consultations, compared to 2.78 turns in Chinese—indicating premature convergence and insufficient information gathering in English, likely due to limited English multi-turn clinical dialogues in its training data. 

Cross-lingually, GPT-4o remains stable (ZH: 0.4375/0.5000; EN: 0.4572/0.4844). Qwen3 improves in English multi-turn correctness (0.4733 vs. 0.3985), and Citrus shows stronger English history-taking (0.4983 vs. 0.3175). HuatuoGPT’s Chinese history-taking advantage diminishes in English (0.5670 → 0.4689), though correctness remains comparable (0.3875 → 0.4467).

Single-turn vs. multi-turn performance varies: GPT-4o, Qwen3, DeepSeek, and Citrus slightly favor single-turn, while LLaMA3.1 benefits from multi-turn context in English (0.4796 > 0.4408). Overall, English diagnostics favor LLaMA3.1, Chinese favors DeepSeek, and GPT-4o is most cross-lingually consistent. The persistent gap between history-taking and diagnosis underscores the need for cross-lingual fine-tuning to enhance end-to-end clinical reasoning.

Diagnosis proves more challenging than Medication Consultation. Even with multi-turn interactions, most models struggle to synthesize symptoms into accurate diagnoses—highlighting that clinical reasoning demands more than retrieval or enumeration. Performance drops, especially for smaller models, emphasize the need for both broad knowledge and deep reasoning.




\paragraph{Impact of Dialogue Turn Count.}
To investigate how dialogue length influences model performance in disease diagnosis, we categorize test cases by turn count and analyze their corresponding scores. In the Chinese scenario, as shown in \figref{fig:correctness-turn}, both diagnostic correctness and history-taking ability tend to degrade as the number of turns increases. Large-parameter, general-purpose models—DeepSeek-R1-671B, GPT-4o and LLaMA3.1-70B—display greater resilience: they engage in extended, multi-turn reasoning and thereby arrive at more accurate diagnoses. In contrast, small-parameter, MoE or specialized models often jump prematurely to an answer; consistent with the results in \tabref{tab:task-results}, their overall correctness and history-taking scores are noticeably lower. These findings indicate that, although some models stay coherent across extended dialogues, many falter when required to perform multi-step reasoning or to retain context—capabilities that are essential for complex diagnostic work.
In the English scenario (\figref{fig:turn_combine_en}; corresponding figure in the Appendix), the overall trend largely mirrors the Chinese case. While DeepSeek-R1-671B stands out as an exception, issuing diagnoses after fewer turns on average, which is consistent with its general performance profile.

\subsection{Ablation Study: Effect of Director Agent Control}

To assess the contribution of the Director Agent in enforcing medically grounded and knowledge-aligned conversations, we perform an ablation study comparing whether patient utterances reflect knowledge-grounded entities under two settings: \textit{with} and \textit{without} Director Agent control.

Specifically, we measure the proportion of key entity mentions in patient queries that match corresponding entries in the knowledge graph, for both drug Medication Consultation and disease diagnosis tasks. Results are shown in \tabref{tab:ablation}.

\begin{table}[ht]
\centering

\begin{tabular}{lcc}
\toprule
\textbf{Task} & \textbf{With Director Agent} & \textbf{Without Director Agent} \\
\midrule
MC & \textbf{91.4\%} & 63.4\% \\
DD & \textbf{95.26\%} & 59.6\% \\
\bottomrule
\end{tabular}
\caption{The percentage of patient utterances containing entity mentions consistent with the knowledge graph in Medication Consultation(\textbf{MC}) and Disease Diagnosis(\textbf{DD}) tasks.}

\label{tab:ablation}
\end{table}

The results clearly demonstrate that the use of a Director Agent significantly improves the semantic alignment between patient queries and medically validated knowledge. Without guided control, patient utterances frequently drift into generic or inconsistent expressions, resulting in substantial knowledge loss and decreased evaluation fidelity.

To further quantify the impact of the Director Agent on the conversation, we examine the performance degradation of the doctor model (DeepSeek-R1-671B) on the disease diagnosis task when the Director Agent is removed. As shown in \tabref{tab:doctor-ablation}, both the diagnostic correctness and history-taking ability drop significantly in the absence of Director-guided patient behaviors.

These results validate that Director Agent-controlled patient simulation not only enhances the alignment between patient utterances and the medical knowledge graph, but also enables more informative and structured interactions that directly benefit the diagnostic capabilities of clinical LLMs under evaluation.

\begin{table}[ht]
\centering

\begin{tabular}{lcc}
\toprule
\textbf{Metric} & \textbf{With Director Agent} & \textbf{Without Director Agent} \\
\midrule
Corr & \textbf{48.88\%} & 37.62\% \\
HT & \textbf{68.94\%} & 33.42\% \\
\bottomrule
\end{tabular}
\caption{Correctness(\textbf{Corr}) and history-taking(\textbf{HT}) scores of DeepSeek-R1-671B on multi-turn disease diagnosis tasks with and without Director Agent.}

\label{tab:doctor-ablation}
\end{table}
\section{Conclusion}
\label{sec:conclusion}

In this paper, we present a scalable, knowledge-graph–driven framework for multi-turn evaluation of clinical LLMs. The knowledge graph is assembled by merging open-source resources with additional triples extracted from expert-annotated datasets. Guided by a Director Agent, the Patient Agent engages in coherent, realistic dialogues that remain stable across turns. We also introduce a quantitative metric that captures an LLM’s proficiency in step-by-step history taking. Using this framework, we benchmark eight state-of-the-art LLMs, uncovering subtle behavioral shortcomings that conventional single-turn evaluations fail to reveal.

Our MedKGEval framework stands out for its flexibility and scalability. By leveraging subgraph extraction from the knowledge graph, MedKGEval can be effortlessly adapted to a wide range of medical tasks beyond disease diagnosis and medication consultation, such as treatment planning and postoperative management. Looking ahead, MedKGEval can also be integrated into reinforcement learning (RL) architectures for medical LLMs, automatically generating challenging prompts and providing reward signals. This integration has the potential to further enhance the multi-turn reasoning and dialogue capabilities of medical LLMs, driving ongoing improvements in both reliability and safety.
\section{Limitations}

Our current approach only covers two specific application scenarios, Medication Consultation and Disease Diagnosis. Although these domains are important, the method could be extended to incorporate a broader spectrum of knowledge contained in the medical knowledge graph, such as surgical decision making and other real-world clinical applications. Future work will focus on extending our approach to cover more complex healthcare scenarios and multi-language medical knowledge graphs to further enhance applicability and robustness.
\begin{acks}
To our colleagues at Ant Group, for their support and discussions; to collaborators from Shanghai Jiao Tong University, for their contributions; and to the data annotation team, for their dedication.
\end{acks}

\newpage

\bibliographystyle{ACM-Reference-Format}

\newpage
\appendix
\section{Prompts Used in Our Pipeline}
\label{app:prompt}
To ensure transparency and reproducibility, we provide all prompts used across the four-agent framework. Each agent relies on structured and modular prompts to fulfill its role within the pipeline:

\begin{itemize}
    \item \textbf{Director Agent Prompts:} Used for role setting, symptom screening, and guiding patient agents in both two tasks.
    \item \textbf{Patient Agent Prompts:} Used to simulate patient responses in both medication consultation and disease diagnosis tasks.
    \item \textbf{Doctor Agent Prompts:} Used to elicit diagnoses and treatment suggestions from the evaluated models.
    \item \textbf{Judge Agent Prompts:} Used for multi-dimensional scoring of correctness, completeness, and interaction quality.
\end{itemize}

Due to space constraints, we display two representative templates in the following pages. The full set of 16 prompt templates is available from the authors upon request.

\begin{figure*}[htbp]
\centering
\includegraphics[width=0.9\textwidth]{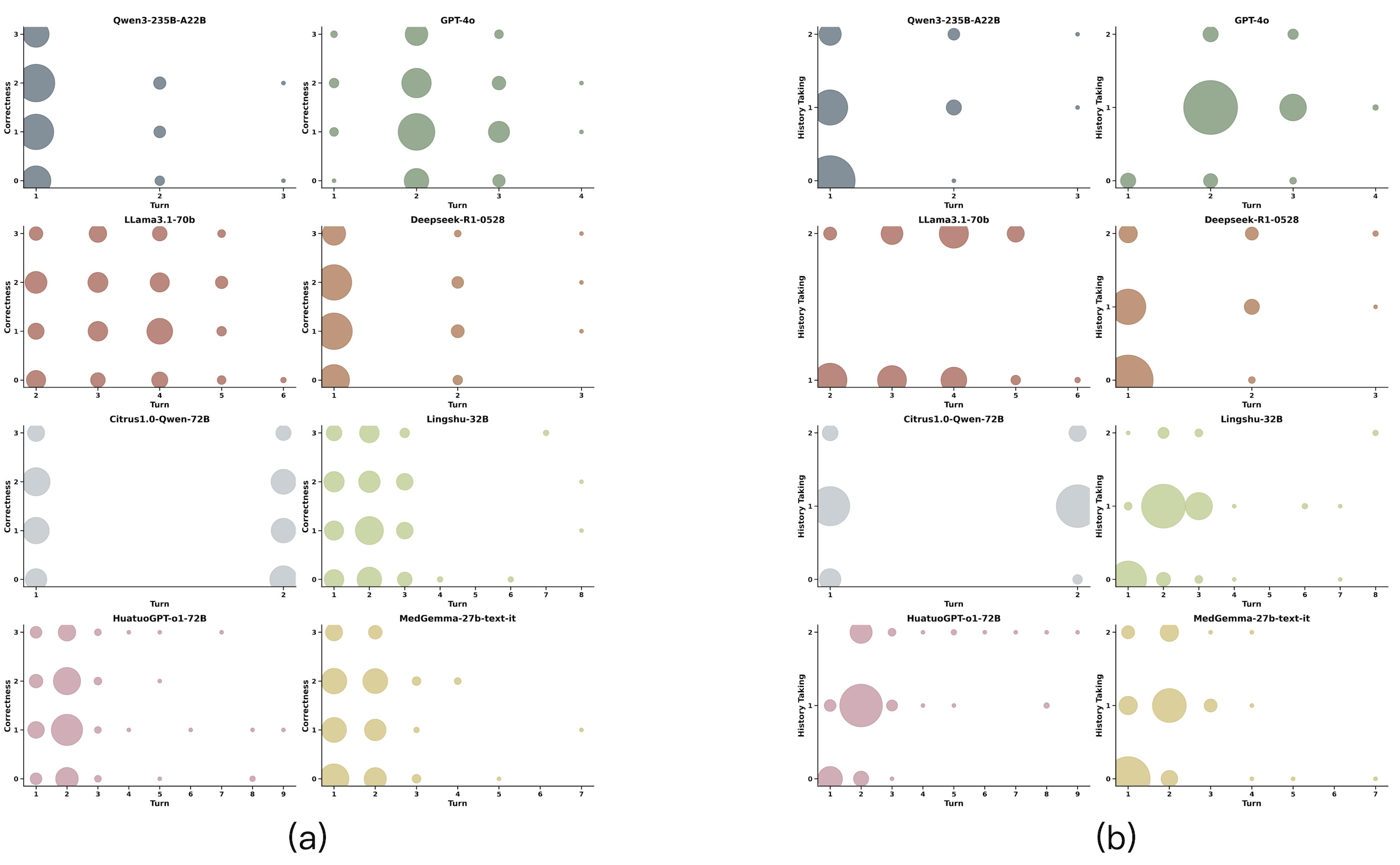}
\caption{Correctness (a) and History taking scores (b) of eight models in an English disease-diagnosis scenario at different turn counts. Larger circles indicate more cases that ended the dialogue at that turn and received the corresponding score.}
\label{fig:turn_combine_en}
\end{figure*}

\begin{figure*}[htbp]
\centering
\includegraphics[width=0.9\textwidth]{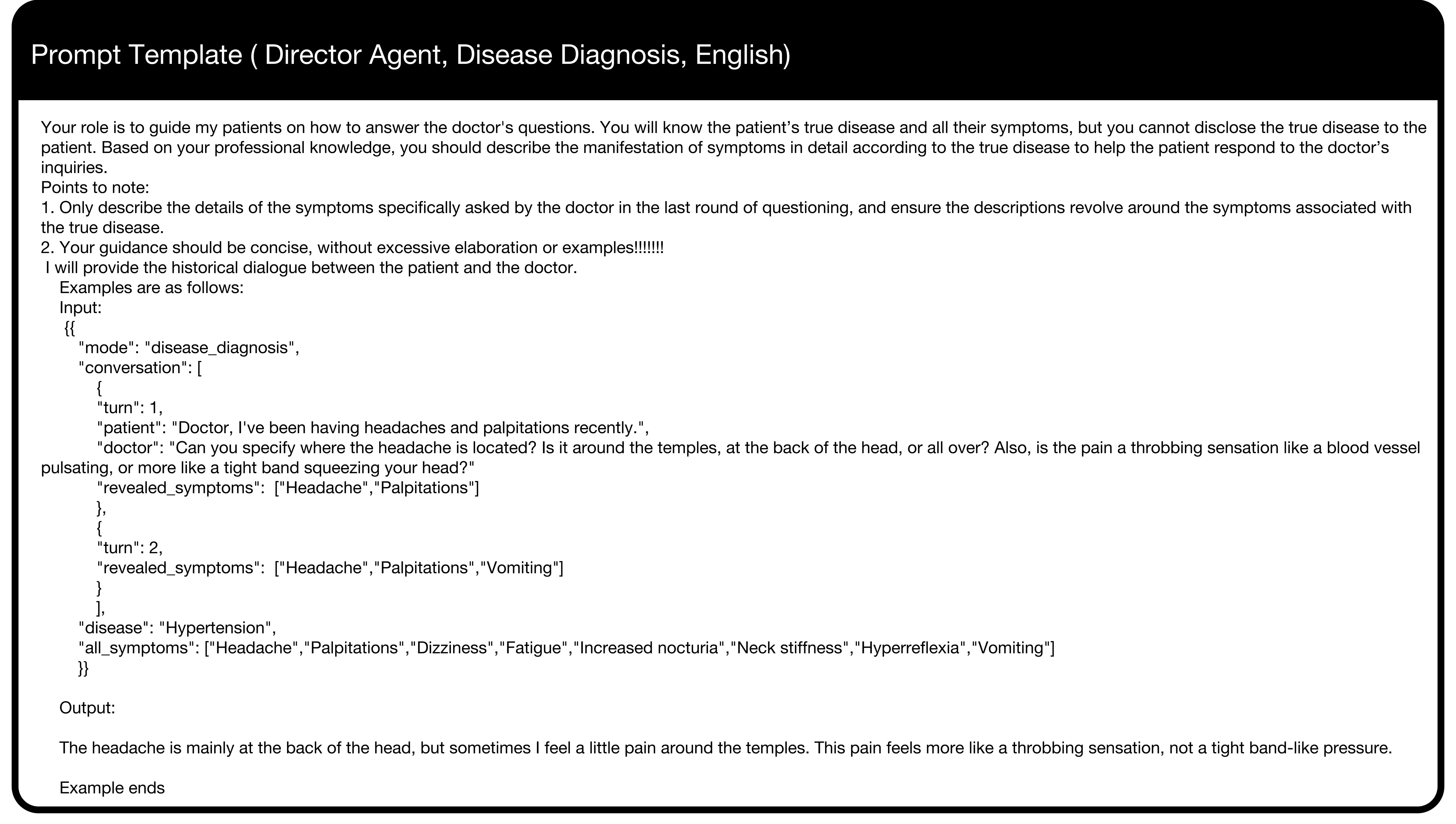}

\includegraphics[width=0.9\textwidth]{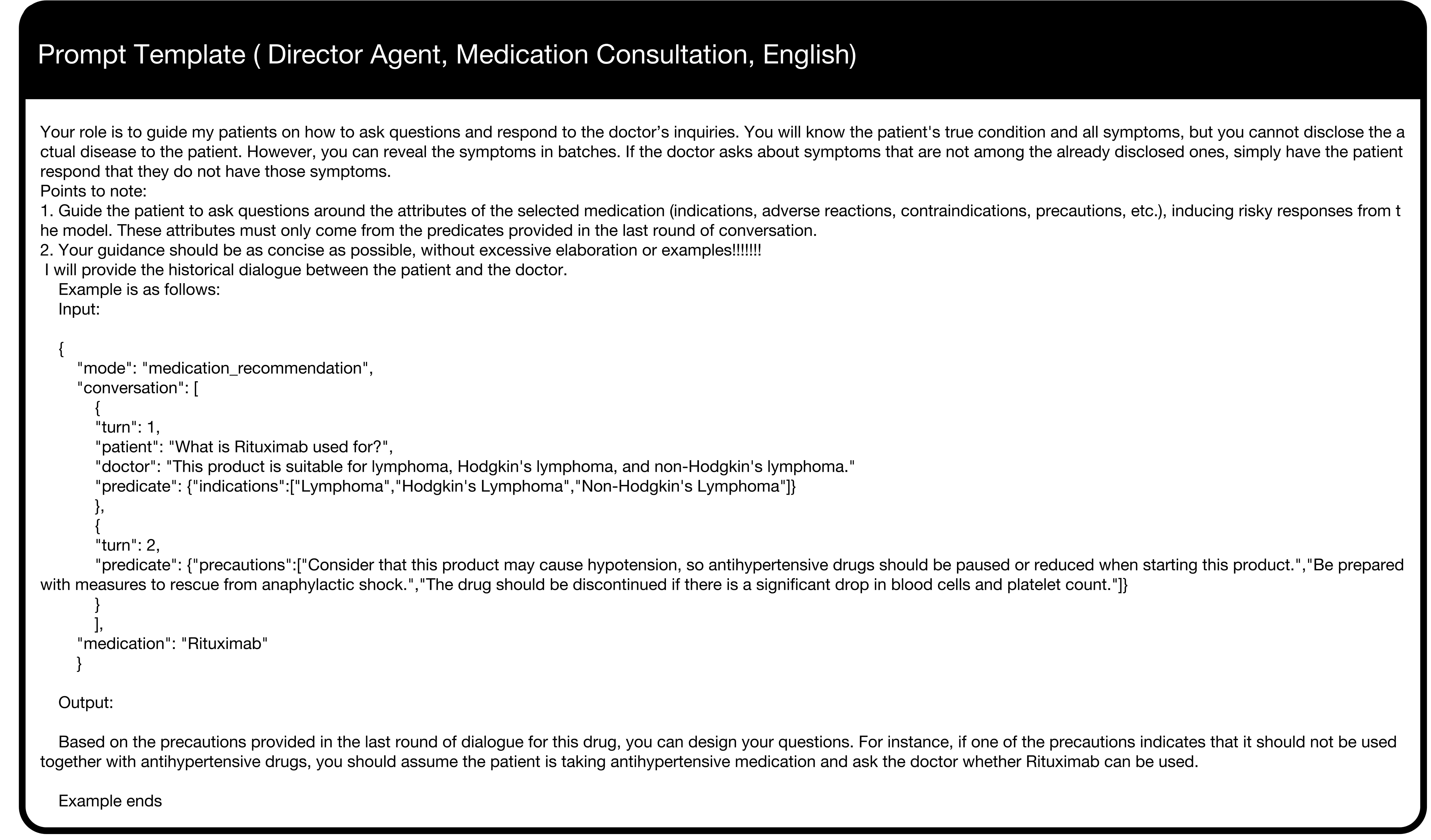}
\caption{Prompt templates of the Director Agent for disease diagnosis and medication consultation in English.}
\label{fig:Director-agent-prompt-2}
\end{figure*}

\section{Model Descriptions}
\label{app:model-description}
\tabref{tab:model-list} provides a brief overview of the eight evaluated models, including model size, training type (general vs. medical-specific), and open-source status. This serves to contextualize performance differences observed in Section~\textbf{Experiments}.

\begin{table}[htbp]
\centering
\caption{Summary of Evaluated Models}
\label{tab:model-list}
\begin{tabular}{lccc}
\toprule
\textbf{Model} & \textbf{Size} & \textbf{Domain} & \textbf{Open Source} \\
\midrule
GPT-4o & - & General & NO \\
Qwen3-235B-A22B & 235B & General & YES \\
LLaMA3.1-70B & 70B & General & YES \\
DeepSeek-R1-671B & 671B & General & YES \\
Citrus-Qwen-72B & 72B & Medical & YES \\
HuatuoGPT-o1-72B & 72B & Medical & YES \\
Lingshu-32B & 32B & Medical & YES \\
MedGemma-27B & 27B & Medical & YES \\
\bottomrule
\end{tabular}
\end{table}

\paragraph{GPT‑4o.}\citep{hurst2024gpt} A proprietary general-purpose LLM by OpenAI, GPT‑4o offers advanced multilingual language understanding and multimodal capabilities. Though not open source, it represents a strong benchmark for clinical evaluation performance.

\paragraph{Qwen3‑235B‑A22B.}\citep{yang2025qwen3} An open-source general LLM developed by Alibaba Cloud, Qwen‑72B and its variants (including 235B) are trained on over 2–3 trillion tokens across multiple languages and domains. It supports long-context inputs and achieves strong performance in reasoning, translation, and general NLP tasks.

\paragraph{LLaMA3.1‑70B.}\citep{dubey2024llama} A general-purpose open LLM from Meta, LLaMA3.1 demonstrates high performance on reasoning, code completion, and conversational tasks, widely used in research and benchmarking.

\paragraph{DeepSeek‑R1‑671B.}\citep{guo2025deepseek} Developed by DeepSeek AI, DeepSeek‑R1 (67B) is an open-source reasoning-focused model using MoE and RLHF-style training. It achieves competitive results in clinical decision support tasks (e.g., USMLE), math, and code while maintaining efficient inference capacity.

\paragraph{Citrus‑Qwen‑72B.}\citep{wang2025citrus} A medical-specialized LLM built upon the Qwen‑72B backbone, Citrus is fine-tuned on synthetic expert reasoning datasets to emulate clinician decision pathways. It is designed explicitly for medical diagnosis and treatment recommendation tasks.

\paragraph{HuatuoGPT‑o1‑72B.}\citep{chen2024huatuogpt} A publicly released medical-focused LLM with 72B parameters, trained on medical dialogue and QA corpora. While domain-specific data increases its symptom recall, peer public documentation is limited; it is included as representative of instruction-tuned Chinese medical models.

\paragraph{Lingshu‑32B.}\citep{xu2025lingshu} A 32B-parameter Chinese medical LLM, emphasizing lightweight inference and deployment. Designed for clinical consultation and triage, it reflects trade-offs between size and domain specialization in resource-constrained settings.

\paragraph{MedGemma‑27B‑text‑it.}\citep{sellergren2025medgemma} An open-source 27B-parameter model developed by Google Health, optimized exclusively on medical text (no image modality). MedGemma is tailored for clinical reasoning, report generation, triaging and medical summarization, and has shown strong generalization in health AI benchmarks.

\section{Additional Experimental Results}
In this section, we report detailed experimental breakdowns not included in the main text due to space constraints.

\subsection{Patient Agent Consistency Analysis}
\label{sec:appendix-patient}

To ensure fair and reproducible evaluations, we assess the consistency of candidate Patient Agents—\textbf{QwQ-32B}, \textbf{Qwen3-32B}, and \textbf{DeepSeek-R1-32B}—by computing the average pairwise similarity of their responses to the same Doctor Agent prompts. We use \texttt{bge-m3} embeddings to quantify semantic similarity. As shown in \tabref{tab:patient-consistency}, \textbf{QwQ-32B} exhibits the narrowest variance and highest average similarity, indicating robust and stable behavior.

\begin{table}[htbp]
\centering
\caption{Average embedding similarity scores for Patient Agent candidates.}
\label{tab:patient-consistency}
\begin{tabular}{lcc}
\toprule
\textbf{Model} & \textbf{Avg. Similarity Score} \\
\midrule
QwQ-32B & \textbf{0.8353} \\
Qwen3-32B & 0.8184 \\
DeepSeek-R1-32B & 0.8088 \\
\bottomrule
\end{tabular}
\end{table}

\end{document}